%% file: main.tex
\documentclass{article} 
\usepackage{iclr2025_conference,times}

\input{math_commands.tex}

\usepackage{hyperref}
\usepackage{url}
\usepackage{multirow}
\usepackage{booktabs}
\usepackage{listings}
\usepackage{amssymb}
\usepackage{pifont}
\newcommand{\cmark}{\ding{51}}%
\newcommand{\xmark}{\ding{55}}%
 \usepackage{graphicx}
 \usepackage{makecell}
\usepackage{float}
\usepackage{colortbl}

\title{Motion-Agent: A Conversational Framework \\ for Human Motion Generation with LLMs}

\renewcommand*{\thefootnote}{\fnsymbol{footnote}}
\author{%
  Qi Wu$^1$\footnotemark{}~, 
  Yubo Zhao$^1$\footnotemark[\value{footnote}]~, 
  Yifan Wang$^1$, 
  Xinhang Liu$^1$, 
  Yu-Wing Tai$^{2}$, 
  Chi-Keung Tang$^{1}$ \\
  \AND
  $^1$The Hong Kong University of Science and Technology\\
  $^2${\bf Dartmouth College}\\
}

\iclrfinalcopy 
\begin{document}

\maketitle
\footnotetext{*Equal contribution.}
\renewcommand*{\thefootnote}{\arabic{footnote}}

\begin{abstract}
While previous approaches to 3D human motion generation have achieved notable success, they often rely on extensive training and are limited to specific tasks. To address these challenges, we introduce {\bf Motion-Agent}, an efficient conversational framework designed for general human motion generation, editing, and understanding. 
Motion-Agent employs an open-source pre-trained language model to develop a generative agent, {\bf MotionLLM}, that bridges the gap between motion and text. This is accomplished by encoding and quantizing motions into discrete tokens that align with the language model's vocabulary. With only 1--3\% of the model's parameters fine-tuned using adapters, MotionLLM delivers performance on par with diffusion models and other transformer-based methods trained from scratch. By integrating MotionLLM with GPT-4 without additional training, Motion-Agent is able to generate highly complex motion sequences through multi-turn conversations, a capability that previous models have struggled to achieve.
Motion-Agent supports a wide range of motion-language tasks, offering versatile capabilities for generating and customizing human motion through interactive conversational exchanges. Project page:~\href{https://knoxzhao.github.io/Motion-Agent}{https://knoxzhao.github.io/Motion-Agent}
\end{abstract}

\input{sections/1_intro}

\input{sections/2_related}

\input{sections/3_method}
\input{sections/4_experiment}
\input{sections/5_discussion}

\bibliography{reference}
\bibliographystyle{iclr2025_conference}

\appendix
\input{sections/appendix}

\end{document}

%% file: math_commands.tex
\usepackage{amsmath,amsfonts,bm}

\def\eqref#1{equation~\ref{#1}}

\def\1{\bm{1}}

\DeclareMathAlphabet{\mathsfit}{\encodingdefault}{\sfdefault}{m}{sl}
\SetMathAlphabet{\mathsfit}{bold}{\encodingdefault}{\sfdefault}{bx}{n}

\DeclareMathOperator*{\argmin}{arg\,min}

%% file: sections/1_intro.tex
\section{Introduction}
Large Language Models (LLMs) have recently attracted much attention in both industry and academia. Many LLMs, such as GPT-4~\citep{achiam2023gpt}, LLaMA~\citep{touvron2023llama}, Gemma~\citep{team2024gemma}, have shown their advanced capabilities, robustness and generalization across various downstream tasks. These progresses have motivated researchers to explore the application of LLMs in multimodal tasks, integrating them with modalities such as images~\citep{koh2024generating}, videos~\cite{zhang2023video}, audio~\citep{audioLM,audiogpt}, and more, resulting in promising outcomes in understanding these different modalities. However, the utilization of LLMs in the context of multimodal {\em generation}, particularly of 3D human motion, remains underexplored, which is crucial for advancing robots and humanoid applications. 

Research in 3D human motion has explored various language-related tasks, including text-conditioned motion generation~\citep{zhang2023generating, Guo_2022_CVPR, MDM, motiondiffuse, Shafir2023priormdm, guo2023momask, jiang2024motiongpt}, motion captioning~\citep{guo2022tm2t,jiang2024motiongpt}, motion reasoning~\citep{endo2023motion,jiang2024motionchain}. 
However, existing methods often require extensive training, leading to high computational demands and inefficiency. 
These models are typically trained on task-specific data, making them data-dependent and limiting their ability to generalize across diverse scenarios. 
They also struggle with handling long, complex prompts with performance degradation. 
Furthermore, most existing models lack the capability to support multi-turn conversational interactions, thus limiting both the generation and refinement processes, and restricting the ability to create dynamic, interactive systems that can seamlessly generate and allow editing motions through dialogue.

Moving forward with the most recent LLM and MLLM development,
in this work, we propose \textbf{Motion-Agent}, a multimodal framework that leverages the generalization and flexibility of pre-trained LLMs. Central to the framework is our new generative agent, \textbf{MotionLLM}, the incorporation of which eliminates the need for extensive pre-training by employing lightweight adapter-based fine-tuning of a pre-trained LLM. Unlike MotionChain~\citep{jiang2024motionchain}, which requires pre-training and large datasets for extensive instruction tuning to achieve conversational control, Motion-Agent integrates MotionLLM with GPT-4 and leverages the LLM's inherent conversational capabilities without additional training. This enables efficient, customizable motion generation, understanding, and multi-turn editing across various tasks.

\begin{figure}[t] 
\vspace{-0.1in}
\centering{
\includegraphics[width=1.0\textwidth]{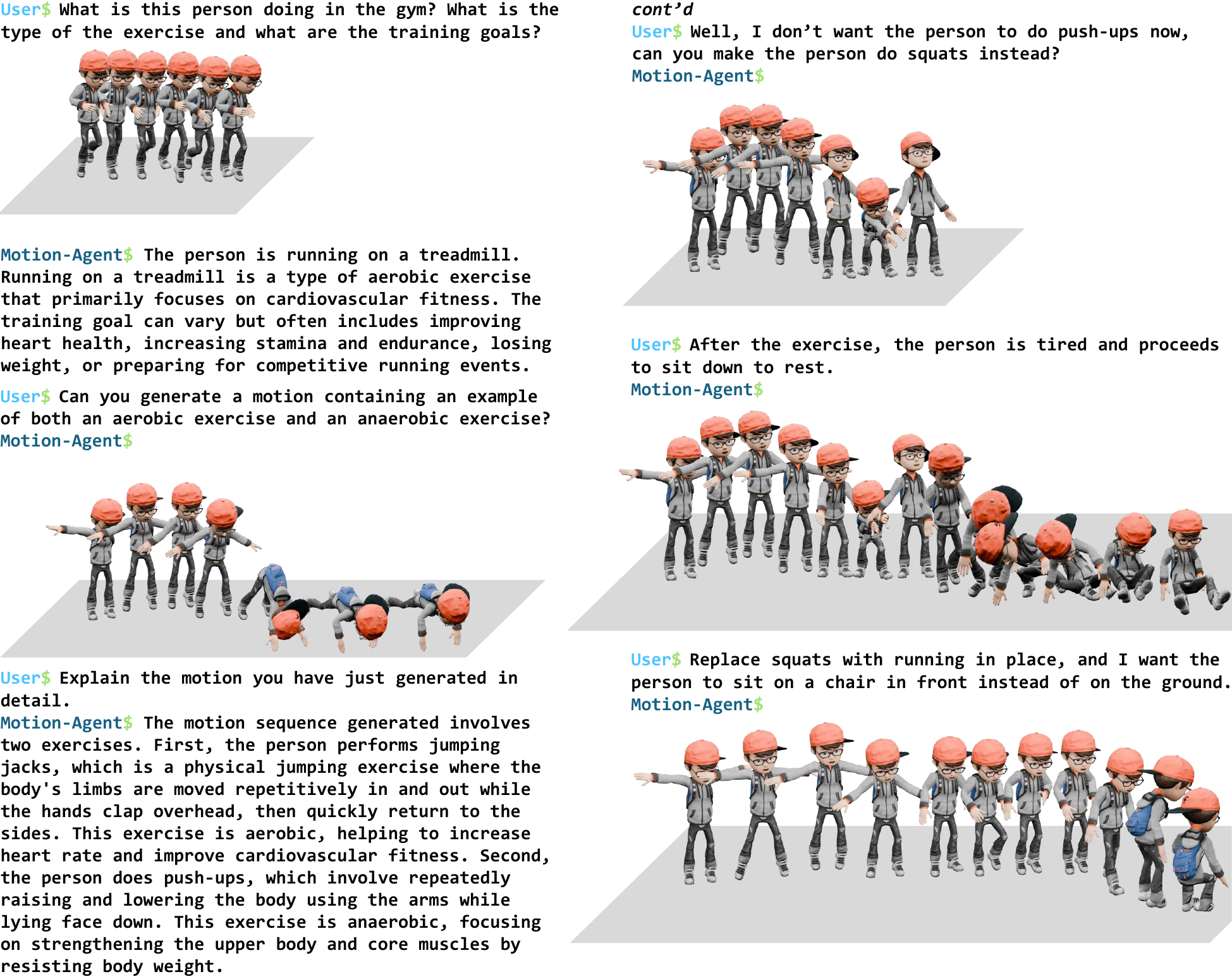}}
\vspace{-0.1in}
\caption{{\bf Multi-turn Conversation Between User and Motion-Agent}. First Turn: Motion Understanding; Second Turn: Motion Generation; Third Turn: Motion Understanding with Previously Generated Motion; Fourth Turn: Motion Editing; Fifth Turn: Continue Motion Generation; Last Turn: Motion Editing on Long Sequence. Note that all turns are continuous. 
}
\vspace{-0.1in}
\label{fig:demo}
\end{figure}

In Motion-Agent, we first train a pair of motion tokenizer and detokenizer. The motion tokenizer encodes motions into motion embeddings and quantizes them into a set of discrete LLM-understandable tokens using a codebook, while the detokenizer reconstructs tokens back to their original continuous forms. This tokenizer-detokenizer pair enables the translation between continuous motion sequences and discrete tokens, facilitating interaction with the LLM while still allowing for the recovery of the original motions from the tokens. MotionLLM is trained by enriching a pre-trained LLM's vocabulary with these additional motion tokens, while keeping the original text tokens unchanged.
Given that motions can be represented as temporal sequences, our tokenization process converts motions into token sequences akin to sentences in natural language. MotionLLM translates between text token sequences and motion token sequences. On top of this, GPT-4 acts as a coordinator, decomposing user instructions to determine the number of calls to MotionLLM and how to structure those calls effectively. The resulting motion token sequences from multiple calls are concatenated and decoded by the detokenizer to produce the final output.

Our Motion-Agent framework leverages pre-trained LLMs in two key ways: (1) fine-tuning a lightweight LLM via adapters to serve as a text-motion translation agent, and (2) using an LLM for conversational interactions without training, thus facilitating multi-turn dialogue for refining generated motions and producing extended motions by iteratively generating and concatenating sequences. Despite training only a small number of parameters, MotionLLM can achieve competitive results in motion generation (text to motion) compared to those trained-from-scratch models with specialized architectures. 
In motion captioning (motion to text), MotionLLM achieves state-of-the-art performances, generating semantically accurate and contextually appropriate text descriptions. MotionLLM enables bidirectional translation between text and motion, outperforming other autoregressive models while using fewer trainable parameters, making it an ideal fit for the overall Motion-Agent framework. By combining MotionLLM with GPT-4, Motion-Agent enables versatile dialogue-based motion generation and reasoning, without requiring specific datasets or extra training for these tasks.

To summarize, our contributions include:
\begin{itemize}
    \item We introduce a simple, efficient conversational framework, Motion-Agent, that utilizes pre-trained LLMs and produces strong results in various motion-language tasks.
    \item We demonstrate the flexibility and versatility of our method by achieving highly customizable motion-language tasks, including long and complex motion generation, multi-turn editing, and multi-turn reasoning.
\end{itemize}

%% file: sections/2_related.tex
\section{Related Work}
\noindent\textbf{Multimodal LLMs}
Recent advancements have integrated large language models (LLMs) with multiple modalities such as image, video, music, audio, and point cloud using different approaches~\citep{liu2023world,han2023onellm,wu2023nextgpt,chen2023xllm,gao2023llamaadapterv2}. Various approaches have been proposed to align different modalities. For instance, Video-LLaMA~\citep{zhang2023video} leverages Q-formers to bridge the gap between modalities. PointLLM~\citep{pointllm} utilizes a projector to align the feature space of point clouds with the feature space of the LLM. VALLE-X \citep{VALL-X} and LlamaGen \citep{sun2024autoregressive} tokenize inputs from various modalities to connect them with language. On the other hand, emerging research~\citep{wu2023visual, lu2024chameleon, du2024compositional} demonstrates promising results with compositional language models. These models, often composed of smaller specialized components, excel in data efficiency and perform well on unseen distributions, aligning with the design of our framework.

\noindent\textbf{3D Human Motion Synthesis}
Modern works can generate human motions based on a variety of inputs such as action labels~\citep{petrovich21actor,Lee2023MultiAct,guo2020action2motion,xu2023actformer}, textual descriptions~\citep{jiang2024motiongpt,fgt2m,zhang2023generating,guo2022tm2t,zhou2023emdm,MDM,tevet2022motionclip,guo2023momask,motiondiffuse,dabral2022mofusion,petrovich22temos,zhang2024motiongpt,pinyoanuntapong2024mmm}, control signals~\citep{omnicontrol,wan2023tlcontrol,petrovich24stmc,como,goel2023iterative}, music or audio~\citep{dabral2022mofusion,tseng2022edge, Zhou_2023_CVPR}, and others~\citep{zhong2024smoodi}. Particularly, text-guided 3D motion generation or text-to-motion has garnered significant interest. Notably, some diffusion models have emerged as powerful tools, such as~\cite{MDM, Shafir2023priormdm,fgt2m,zhou2023emdm,omnicontrol,motiondiffuse}. Despite the proficiency in generating motions, diffusion models necessitate manual length control of the generated motions with limited flexibility. In addition to diffusion models, which employ continuous motion representation, discrete token-based methods utilizing Vector Quantized Variational Autoencoders (VQ-VAEs) have also demonstrated promising results. Notable examples include TM2T~\citep{guo2022tm2t}, T2M-GPT~\citep{zhang2023generating}, MotionGPT~\citep{jiang2024motiongpt} and MoMask~\citep{guo2023momask}. Most existing works in both approaches focus on conditional generation to translate between modalities. In our work, we emphasize generating human motion through complex, customized user conversations while proposing a training-efficient approach to bridge these modalities using pre-trained LLMs.

\noindent\textbf{Conversational Control For Human Motion}
Generating 3D human motion through conversation is more flexible, which allows users to customize versatile requests and control motion via iterative refinement. While models like MotionGPT~\citep{jiang2024motiongpt} handle some simple single-turn tasks using instruction tuning, and MotionChain~\citep{jiang2024motionchain} supports multi-turn interactions by sampling single-turn data into multi-turn training data, both methods rely heavily on extensive instruction tuning and additional data. In contrast, our Motion-Agent framework uses a composition of LLMs to eliminate extra training. By training the translation agent, MotionLLM, solely on the original text-motion paired data, our method eliminates the need for further data or training, resulting in higher efficiency and broader generalizability.

%% file: sections/3_method.tex
\begin{figure}[t] 
\vspace{-0.1in}
\centering{
\includegraphics[width=\textwidth]{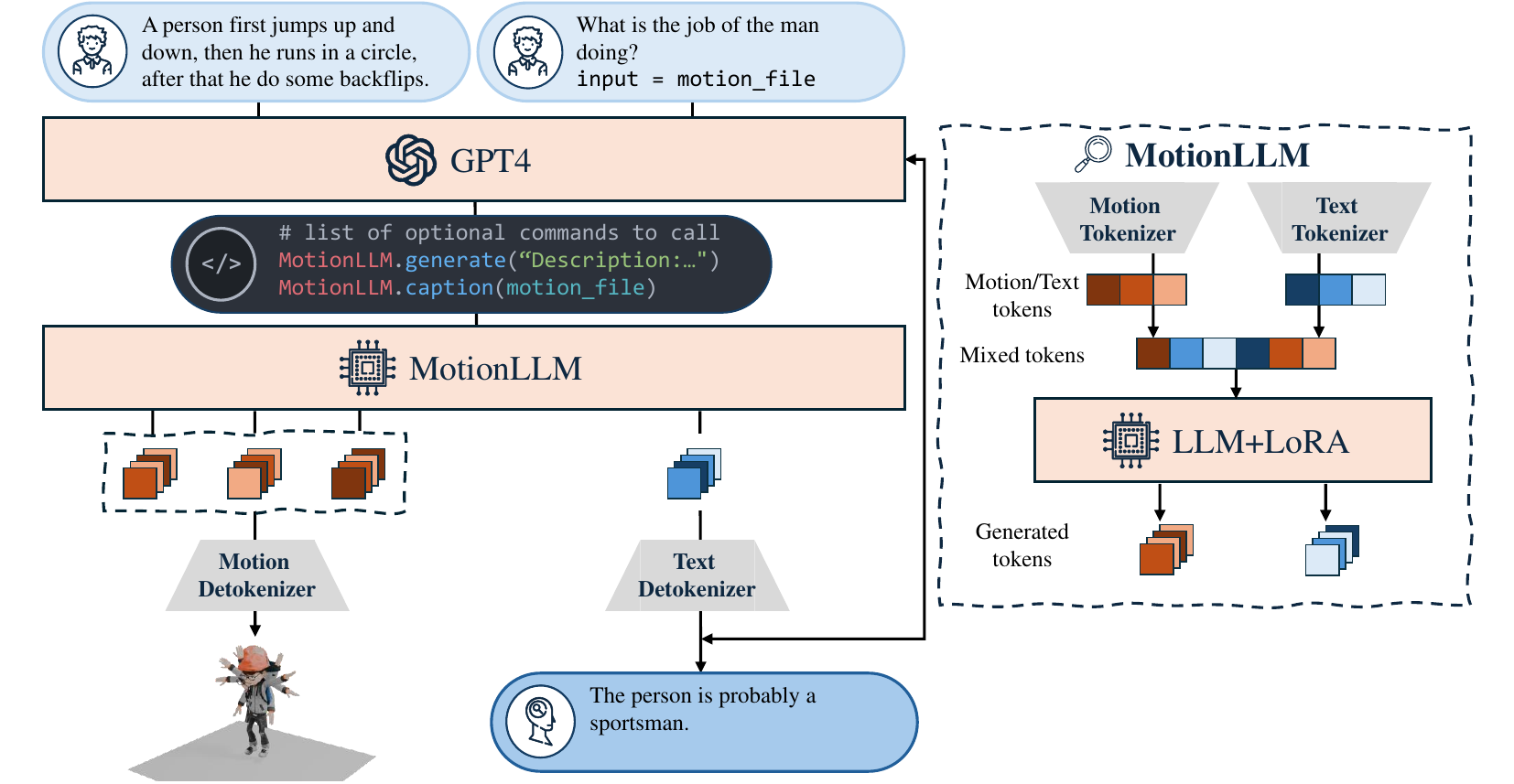}}
\vspace{-0.2in}
\caption{{\bf Motion-Agent pipeline}. GPT-4 can interact with the translation agent (i.e., MotionLLM) to generate or interpret motions based on input requirements. The generated motion tokens are concatenated and decoded, and the textual caption produced by MotionLLM is returned and processed by GPT-4.}
\vspace{-0.1in}
\label{fig:pipeline}
\end{figure}

\section{Method}
As shown in Fig.~\ref{fig:pipeline}, our Motion-Agent framework primarily consists of three components: an LLM (i.e., GPT-4) for conversational interaction and prompting control, a pair of motion tokenizer/detokenizer, and a translation agent (i.e., MotionLLM). The text tokenizer is inherited from the LLMs and remains unchanged, while the motion tokenizer and detokenizer are trained together to ensure proper reconstruction of motion sequences. Once trained, the motion tokenizer and detokenizer are kept fixed. The motion detokenizer plays a key role in smoothing the transitions between different motion sequences, ensuring seamless integration of motion outputs.

To ensure bidirectional understanding, our framework also enables motion comprehension. Thus, the agent should also be capable of generating textual captions from given motions upon request. This bidirectional translation is crucial for applications such as answering questions about motions or generating descriptions, where models that can only perform motion generation are not suitable. On the other hand, our proposed MotionLLM can indeed be a good fit, ensuring bidirectional translation within a unified architecture. 

\subsection{The Motion-Agent Framework} 
In this framework, GPT-4 serves as the coordinator of both motion generation and comprehension, enabling seamless interaction between users and a multimodal text-motion agent. The agent is responsible for translating between text and motion modalities. 
Within the conversation, the input to GPT-4 consists of two components: a fixed instruction prompt $p$, which provides guidelines for interacting with the text-motion agent, and a customized request $c$ from the user. 
Based on this input, GPT-4 generates a structured plan, determining whether the agent should perform tasks such as motion generation or captioning. It also decides how many times to invoke the agent and specifies the arguments for each invocation. This plan, formatted as a JSON file, is then parsed and executed by the agent to carry out the specified tasks.

For generation, the agent generates motion token sequences corresponding to each set of arguments, and these sequences are concatenated for universal decoding. Specifically, let $G$ represent the agent and $[\textbf{a}_i]_{i=1}^N$ the arguments for each of the $N$ calls determined by GPT-4. The resulting motion token sequences $\textbf{z}_i = G(\textbf{a}_i)$ are concatenated to form a single sequence $\textbf{z} = (\textbf{z}_1, \textbf{z}_2, \dots, \textbf{z}_N)$. This sequence is decoded by the decoder $D$ to produce the final motion, $\textbf{m} = D(\textbf{z})$, as will be outlined in the tokenization section.

For motion understanding and reasoning, the agent generates textual captions of the motions, which are then returned to GPT-4. This allows GPT-4 to interpret the motion and respond to user queries accordingly, enabling seamless interactions between users and the system through both motion generation and comprehension.

Since LLMs such as GPT-4 possess strong multi-turn conversational abilities, users can continuously ask the model to refine, edit, or extend previous generations, as well as pose additional questions. In response, GPT-4 will re-generate the plan or provide answers, thus providing an interactive and adaptive system. This dynamic interaction leads to a unified framework that supports an exceptionally wide variety of combinations, lengths, and task complexities, offering enhanced flexibility and customization across both motion generation and comprehension.

\subsection{Motion Tokenization}
In order to align better with LLM's next-token prediction mechanism, we tokenize motions into discrete representations using Vector Quantization (VQ) and Variation AutoEncoders (VAE). This VQ-VAE approach is widely adopted by \citet{guo2022tm2t}, \citet{zhang2023generating}, \citet{jiang2024motiongpt}, and \citet{guo2023momask}.

In our motion tokenization, a motion sequence is represented as $\textbf{m}_{1:T}\in \mathbb{R}^{T\times D}$ and is first encoded using an encoder $E$ to motion embeddings $\textbf{z}_{1:T/N}\in \mathbb{R}^{T/N\times d}$, where $N$ is the downsampling rate and $d$ is the number of the hidden dimensions. Then the motion embeddings are quantized by a quantizer using a codebook $\textbf{C}=\{\textbf{c}_k\}_1^K$, where $K$ is the codebook size and each $\textbf{c}_k\in \mathbb{R}^d$. The quantization results can be represented as $\hat{\textbf{z}}_{1:T/N}$, where
\begin{equation*}
    \hat{\textbf{z}_t} = \argmin_{\textbf{c}_k\in \textbf{C}} ||\textbf{z}_t - \textbf{c}_{k}||_2
\end{equation*}
The original sequence can be reconstructed by the decoder $D$: $\hat{\textbf{m}}_{1:T} = D(\hat{\textbf{z}}_{1:T/N})$. 

We follow \citet{zhang2023generating} to optimize the VQ-VAE, using reconstruction loss together with a commitment loss. We also add an additional regularization on the joint positions $\textbf{p}$ to enhance the generation performance. The loss can be formulated as:
\begin{equation*}
    \mathcal{L}_{vq}=\underbrace{||\textbf{m}-\hat{\textbf{m}}||_1}_{\mathcal{L}_{re}} + \alpha\underbrace{||\textbf{p}-\hat{\textbf{p}}||_1}_{\mathcal{L}_{p}} + \beta\underbrace{ ||\textbf{z}-sg[\hat{\textbf{z}}]||_2}_{\mathcal{L}_{\it{commit}}}
\end{equation*}
where $sg[\cdot]$ is the stop-gradient operation, $\alpha$ and $\beta$ are weighting factors. The codebooks are trained using exponential moving average (EMA) and codebooks reset following T2M-GPT~\citep{zhang2023generating}. After training, the tokenizers are frozen for further usage.

\subsection{LLM-based Motion-Language Agent}
\label{subsec:method_mm_motion_language}
Following tokenization, the motion representation is discretized into $K$ distinct motion tokens. We utilize the indices of these motion tokens from the codebook to construct the motion token vocabulary $\textbf{V}_m = \{{\tt<Motion\_i>}\}_{i=1}^K$. In addition, we introduce special tokens ``{\tt <Motion>}" and ``{\tt </Motion>}" to denote the start and end of a motion token sequence. These special tokens, together with the motion tokens, form a new vocabulary set $\textbf{V}_M$ of size $K+2$. This vocabulary will then be appended to the pre-trained LLM's vocabulary. 

After expanding the LLM's vocabulary, a motion can now be denoted as a token sequence that is understandable by the LLM. During the generation process, the LLM predicts the succeeding token by maximizing the probability $p_{\theta}(x_t|x_{<t},c)$, where $x_{1:T}$ is the target token sequence and $c$ represents the prompt. This prediction is performed iteratively in an autoregressive manner. Consequently, the training objective aims to maximize the log-likelihood $\mathcal{L}_{\mathit{LLM}}=-\sum \log p_{\theta}({x_t|x_{x_{<t}},c})$. 

During the inference process, our approach utilizes instructive prompts such as ``Generate a motion that matches the following input human motion description.'' accompanied by a sentence describing the desired motion. The LLM then proceeds to predict tokens autoregressively until it predicts the ``{\tt </Motion>}" token, indicating the completion of the motion generation. This autoregressive process allows for the generation of motions with variable lengths, adapting to the specific requirements of the given description.

To fine-tune the LLM, we employ LoRA~\citep{hu2021lora}. Throughout the whole training process, the tokenizer, the embeddings, and the output layer of the original text tokens remain unchanged and frozen. Only the additional adapters are trained. These LoRA adapters are trained for the task at hand (generation or captioning) while maintaining a general architecture where multiple adapters can coexist harmoniously. This approach allows us to leverage the power of LLMs while tailoring them to specific motion-language tasks, ensuring efficient and effective training without altering the core components of the LLM.

%% file: sections/4_experiment.tex
\section{Experiments}
We assess our Motion-Agent framework with general and complex conversational user inputs, demonstrating its ability to handle intricate, multi-turn interactions. We also evaluate MotionLLM on single-turn motion generation and motion captioning tasks.

\begin{table}[t]
\resizebox{\textwidth}{!}{
\begin{tabular}{cccccc}
\toprule
                Methods & Motion Generation & Captioning & Multi-turn Editing & Reasoning & Composition\\ \midrule
MotionGPT~\citep{jiang2024motiongpt} & short     & \cmark    & \xmark    & \xmark    & \xmark    \\
MoMask~\citep{guo2023momask} & short     & \xmark    & \xmark    & \xmark    & \xmark    \\
\em{MotionChain}~\citep{jiang2024motionchain} & short     & \cmark    & \cmark    & \cmark    & \cmark    \\
Ours       & long      & \cmark    & \cmark    & \cmark    & \cmark    \\
\bottomrule
\end{tabular}
}\vspace{-0.1in}
\caption{Comparison on functionalities among recent motion generation models. Italicized \emph{model} indicates the corresponding model requires pre-training and task-specific tuning.}\vspace{-0.15in}
\label{tab:motion-agent-compare}
\end{table}

\subsection{Experiment Setup}
\noindent\textbf{Datasets.}
Our experiments on MotionLLM are conducted with KIT Motion Language Dataset (KIT-ML)~\citep{Plappert2016}, HumanML3D~\citep{Guo_2022_CVPR}. KIT-ML contains 3,911 human motion sequences, while HumanML3D dataset, obtained from AMASS~\citep{AMASS:ICCV:2019} and HumanAct12~\citep{guo2020action2motion}, contains 14,616 human motions sequences with 44,970 textual descriptions. For Motion-Agent, we use the MotionLLM model which is trained on HumanML3D.

\vspace{2mm}
\noindent\textbf{Evaluation Metric.} 
For motion generation, we follow T2M~\citep{Guo_2022_CVPR}. Global representations of motion and textual descriptions are first extracted with the pre-trained network in~\citep{Guo_2022_CVPR} and then measured in the following: 1) Text matching: R-precision (Top-1, Top-2, and Top-3 accuracy) by ranking Euclidean distances between motion and text embeddings, and MM Dist, which measures the average distance between text and generated motion embeddings. 2) Generation diversity: quantifies the variance of generated motions across all descriptions. 3) Motion fidelity: FID assesses the distance between the distribution of real and generated motions, reflecting how closely they match real motion distributions. For motion captioning, we follow TM2T~\citep{guo2022tm2t} to evaluate the quality of motion captioning by facilitating linguistic metrics from natural language studies, including Bleu~\citep{bleu}, Rouge~\citep{lin-2004-rouge}, Cider~\citep{vedantam2015cider}, and Bert Score~\citep{zhang2020bertscore}. 

\vspace{2mm}
\noindent\textbf{Implementation Details.}
We utilize GPT-4~\citep{achiam2023gpt} as the conversational LLM in our Motion-Agent framework, which offers enhanced textual control and interaction capabilities.
In our tokenizer, we set the downsampling rate $N$ to 4, the hidden dimension $d$ to 512, and the codebook size $K$ to 512. The weighting factors $\alpha$ and $\beta$ for $\mathcal{L}_{p}$ and $\mathcal{L}_{\it{commit}}$ are set to 0.5 and 0.02 respectively. For MotionLLM, we employ Gemma2-2b-it~\citep{team2024gemma2}, a lightweight open-source LLM from Google, which offers accessibility and can be deployed on a single consumer-level GPU. The LoRA rank is set to 64 for generation and 32 for captioning, the values of alpha remain the same with the rank. All of our experiments are conducted on NVIDIA RTX4090s.

\subsection{Results of Motion-Agent}

\begin{figure}[t] 
\vspace{-0.1in}
\centering{
\includegraphics[width=\textwidth]{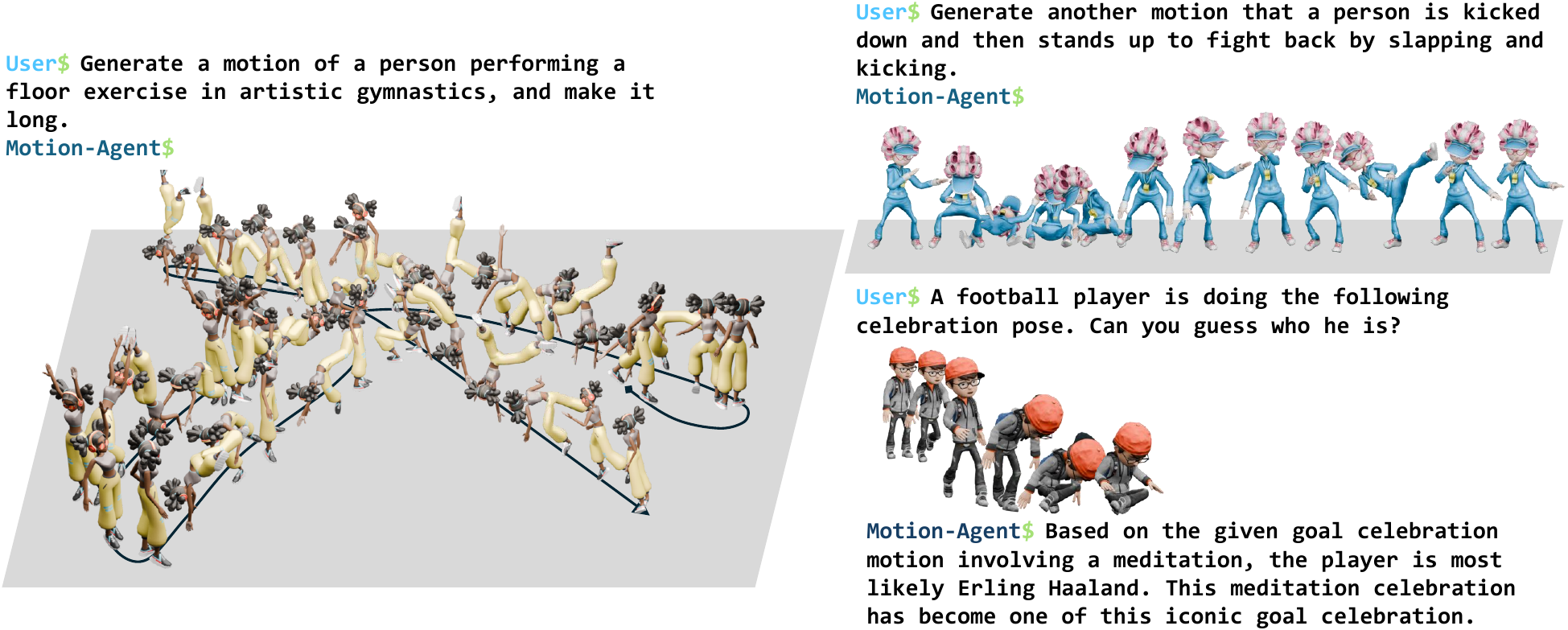}}
\vspace{-0.2in}
\caption{Motion-Agent can comprehend abstract, complex user prompts and generate accurate, long motions. It also understands and answers user questions based on real-world knowledge. Notably, the three turns in this figure stem from a continuous conversation, demonstrating the flexibility of its multi-turn capability in scenarios that should not be influenced by previous turns.}
\label{fig:demo_result}
\end{figure}

\begin{figure}[t] 
\centering{
\includegraphics[width=\textwidth]{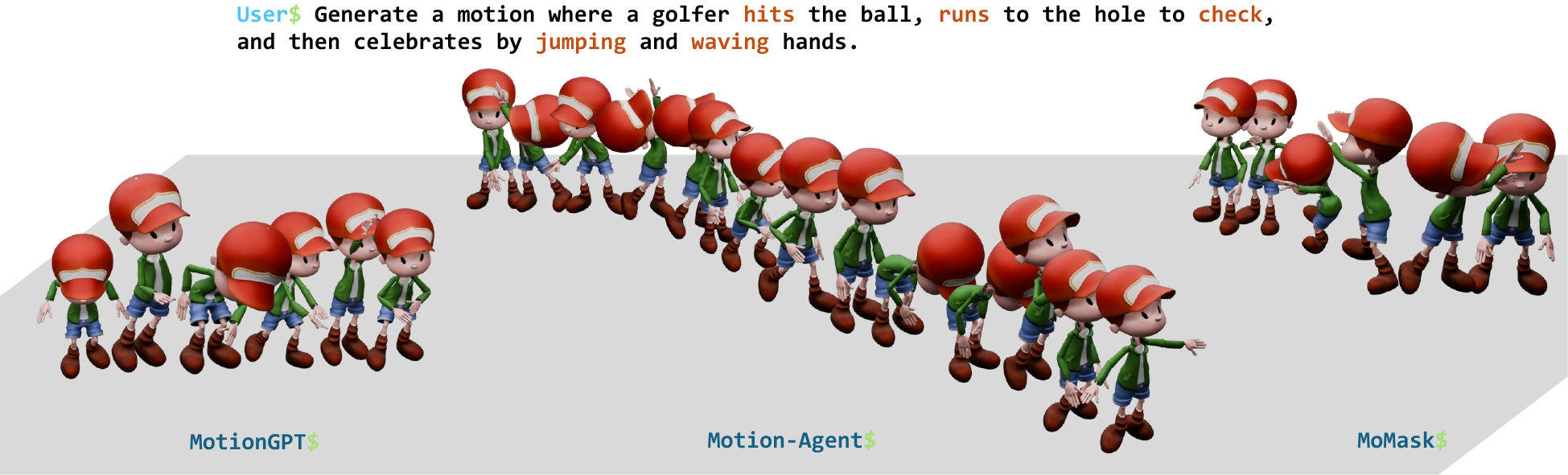}}
\vspace{-0.2in}
\caption{{\bf Comparison with Other Methods}. Our Motion-Agent accurately generates motions involving a series of actions, while other models struggle with more complex descriptions like this, resulting in short and unclear motions.}
\vspace{-0.1in}
\label{fig:comparison}
\end{figure}

In this section, we present the results of our Motion-Agent framework, demonstrating its ability to generate long outputs through complex combinations of tasks via multi-turn conversations. It is important to note that no established ground truth exists for such tasks, aside from text-motion translation, where we do not conduct additional training for these extended tasks.

As shown in Table~\ref{tab:motion-agent-compare}, Motion-Agent is proficient in various motion-language tasks, generating long motion sequences through natural conversational user interactions.
MotionGPT~\citep{jiang2024motiongpt} supports bidirectional translation but lacks versatility, while MoMask~\citep{guo2023momask} excels in generation but is limited to this task. Although MotionChain~\citep{jiang2024motionchain} 
can perform similar functions, it requires additional datasets for task-specific instruction tuning. 
These methods, along with most existing approaches, are restricted to relatively short motion sequences.
In contrast, without training on additional datasets, our Motion-Agent can generate longer sequences, accurately matching the given prompts, as indicated in Figure~\ref{fig:comparison}. 
While HumanML3D~\citep{Guo_2022_CVPR} contains a wide range of human motions, its sequences are generally short and atomic, lasting less than 10 seconds. By decomposing descriptions of long motions into a series of short motions using LLMs and subsequently concatenating these short motions into longer sequences, our Motion-Agent can theoretically achieve infinite motion generation. This decompose-and-integrate approach can thoroughly leverage existing data for long motion generation, mapping known data distributions to unknown ones and enhancing both efficiency and scalability.

\begin{figure}[t] 
\centering{
\includegraphics[width=\textwidth]{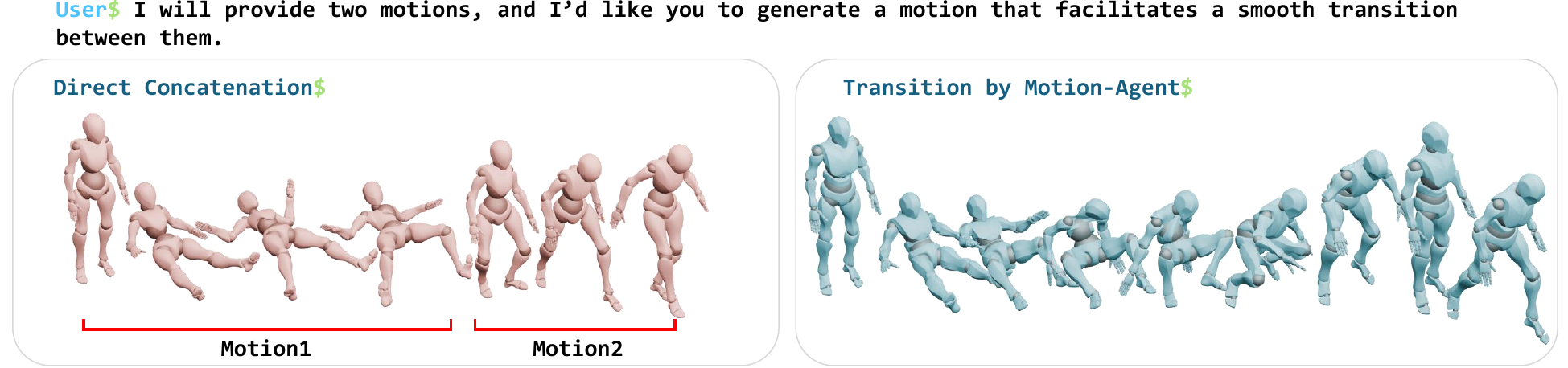}}
\vspace{-0.2in}
\caption{Motion-Agent can compose motions with smooth transitions. In this example, the two motions ``a person falls down on the back'' and ``a person is walking'' are provided to Motion-Agent in two turns. The system then generates a ``stand up'' motion to facilitate a seamless composition of the two motions.}
\vspace{-0.1in}
\label{fig:tran}
\end{figure}

The integration of LLMs also improves the system's ability to interpret vague, abstract, or complex motion descriptions, allowing for iterative refinement through multi-turn conversations.
Figure~\ref{fig:demo} already illustrates our framework's strong multi-turn contextual capabilities, enabling it to understand, extend, and edit the results of previous turns effectively. Additionally, our multi-turn functionality facilitates non-contextual requests, as evidenced by the results in Figure~\ref{fig:demo_result}, which were generated within a single conversation comprising multiple turns. This flexibility allows users to avoid restarting for new requests. Furthermore, the results in Figure~\ref{fig:demo_result} demonstrate that our method can accommodate general, customized, and complex user requests through conversational and iterative exchanges. Our Motion-Agent is also capable of generating transition motions to connect and compose movements seamlessly, as shown in Figure~\ref{fig:tran}, an ability that previous motion generation models struggled to achieve. This further demonstrates the motion understanding and generation capabilities of our method.

More qualitative results are presented in the appendix~\ref{subsubsec:moton-agent-qualitative} and the supplementary material.

\subsection{Evaluations of MotionLLM}
We evaluate MotionLLM on both text-to-motion and motion-to-text tasks to validate that it achieves satisfactory results. MotionLLM is focused on enabling bidirectional translation with minimal training load, while still maintaining competitive performance across key benchmarks. Quantitative results are shown in Table~\ref{tab:t2m_table}.

For generation, we compare our model with state-of-the-art (SOTA) approaches, including diffusion models~\citep{MDM, chen2023executing, motiondiffuse} and token-based models~\citep{zhang2023generating, jiang2024motiongpt, guo2023momask}. Despite {\it fine-tuning only a small number of parameters}, our model performs competitively against these models {\it trained from scratch}. This demonstrates our advantages of leveraging the generalization and robustness capabilities of LLMs. Additionally, our model exhibits low MMDist, high R Precision and high Diversity, indicating strong motion-language understanding and generative capabilities. Note that MoMask~\citep{guo2023momask} and the diffusion models are non-autoregressive, requiring known target lengths for generation, and evaluate using ground truth lengths. However, since the FID metric measures the distance between the distribution of generated results and ground truth, variable length generated by autoregressive models can lead to higher FID scores. Yet, our MotionLLM achieves a lower FID than some other autoregressive models such as MotionGPT~\citep{jiang2024motiongpt} with only about one-third of trainable parameters.
Additionally, the autoregressive nature of our model offers advantages over non-autoregressive models when ground truth motion lengths are not provided. This makes MotionLLM a better fit for our Motion-Agent framework, as it eliminates the need for specifying motion lengths. 

In Sec.~\ref{subsec:ablation}, we provide further analysis and evidence that increasing the model size can lead to overall improvements in performance scores. For a more economical choice, we selected one of the smallest LLMs (Gemma2-2B) available to the public.

For captioning, we compare models capable of bidirectional generation. Leveraging the strong text processing capabilities of LLMs, MotionLLM produces accurate descriptions of human motions. We assess the generated captions using linguistic metrics from \citet{guo2022tm2t}, which calculate semantic similarities to ground truth captions. To ensure an accurate evaluation, we follow \citet{jiang2024motiongpt} by using the unprocessed ground truth texts, as \citet{guo2022tm2t} ignores grammatical tense and plural forms. As demonstrated in Tab.~\ref{tab:t2m_table}, our method outperforms previous SOTA approaches across all metrics by a large margin, thanks to the language abilities of pre-trained LLMs.

\begin{table}[t]
	\centering
	\resizebox{\textwidth}{!}{
\begin{tabular}{ccccccc}
\toprule
\multirow{2}{*}{Tasks}               & \multirow{2}{*}{Methods} & \multicolumn{2}{c}{R Precision $\uparrow$}                  & \multirow{2}{*}{FID $\downarrow$} & \multirow{2}{*}{MultiModal Dist $\downarrow$} & \multicolumn{1}{c}{\multirow{2}{*}{Diversity$\uparrow$}} \\ \cline{3-4}
                                    &                          & Top 1                        & Top 3                        &                                   &                                               & \multicolumn{1}{c}{}                                     \\ \midrule
\multirow{10}{*}{Generation} & T2M~\citep{Guo_2022_CVPR}                      & $0.457^{ \pm .002}$          & $0.740^{ \pm .003}$          & $1.067^{ \pm .002}$               & $3.340^{ \pm .008}$                           & \multicolumn{1}{c}{$9.188^{ \pm .002}$}                  \\
                                    & TM2T~\citep{guo2022tm2t}                     & $0.424^{ \pm .003}$          & $0.729^{ \pm .002}$          & $1.501^{ \pm .017}$               & $3.467^{ \pm .011}$                           & \multicolumn{1}{c}{$8.589^{ \pm .076}$}                  \\
                                    & \textit{MDM}~\citep{MDM}                      & $0.320^{ \pm .005}$          & $0.611^{ \pm .007}$          & $0.544^{ \pm .044}$               & $5.566^{ \pm .027}$                           & \multicolumn{1}{c}{$9.559^{ \pm .086}$}                  \\
                                    & \textit{MLD}~\citep{chen2023executing}                      & $0.481^{ \pm .003}$          & $0.772^{ \pm .002}$          & $0.473^{ \pm .013}$               & $3.196^{ \pm .010}$                           & \multicolumn{1}{c}{$9.724^{ \pm .082}$}                  \\
                                    & \textit{MotionDiffuse}~\citep{motiondiffuse}            & $0.491^{ \pm .001}$          & $0.782^{ \pm .001}$          & $0.630^{ \pm .001}$               & $3.113^{ \pm .001}$                           & \multicolumn{1}{c}{$9.410^{ \pm .049}$}                  \\
                                    & T2M-GPT~\citep{zhang2023generating}                  & $0.491^{ \pm .003}$          & $0.775^{ \pm .002}$          & $\underline{0.116}^{ \pm .004}$               & $3.118^{ \pm .011}$                           & \multicolumn{1}{c}{$\underline{9.761}^{ \pm .081}$}                  \\
                                    & MotionGPT~\citep{jiang2024motiongpt}                & $0.492^{ \pm .003}$          & $0.778^{ \pm .002}$          & $0.232^{ \pm .008}$               & $3.096^{ \pm .008}$                           & \multicolumn{1}{c}{$9.528^{ \pm .071}$}                  \\
                                    & MotionChain~\citep{jiang2024motionchain}              & $0.504^{\pm .003}$           & $0.790^{\pm .003}$           & $0.248^{\pm .009}$                & $3.033^{\pm .010}$                            & \multicolumn{1}{c}{$9.470^{\pm .075}$}                   \\
                                    & \textit{MoMask}~\cite{guo2023momask}                   & $\textbf{0.521}^{ \pm .002}$ & $\textbf{0.807}^{ \pm .002}$ & $\textbf{0.045}^{ \pm .002}$      & $\textbf{2.958}^{ \pm .008}$                  & \multicolumn{1}{c}{${9.620}^{ \pm .064}$}                \\
                                    \cline{2-7}
                                    & \textbf{MotionLLM}                & $\underline{0.515}^{ \pm .004}$          & $\underline{0.801}^{ \pm .004}$          & $0.230^{ \pm .009}$               & $\underline{2.967}^{ \pm .020}$                           & \multicolumn{1}{c}{$\textbf{9.908}^{ \pm .102}$}         \\ \midrule 
\multirow{5}{*}{Captioning}  &                          & Bleu@1$\uparrow$             & Bleu@4$\uparrow$             & Rouge$\uparrow$                   & Cider$\uparrow$                               & Bert Score$\uparrow$                                      \\ \cline{2-7} 
                                    & TM2T~\citep{guo2022tm2t}                     & \underline{48.90}                        & 8.27                         & \underline{38.1}                              & 15.80                                         & 32.2                                                      \\
                                    & MotionGPT~\citep{jiang2024motiongpt}                & 48.20                        & 12.47                        & 37.4                              & 29.20                                         & 32.4                                                      \\
                                    & MotionChain~\citep{jiang2024motionchain}              & 48.10                        & \underline{12.56}                        & 33.9                              & \underline{33.70}                                         & \underline{36.9}                                                      \\
                                    \cline{2-7}
                                    & \textbf{MotionLLM}                & $\textbf{54.53}$             & $\textbf{17.65}$             & $\textbf{48.7}$                   & $\textbf{33.74}$                              & $\textbf{42.63}$                                           \\ \bottomrule
\end{tabular}
    }
    \vspace{-0.1in}
    \caption{\textbf{Quantitative evaluation of MotionLLM on the HumanML3D~\citep{Guo_2022_CVPR} test set}. For motion generation, we follow T2M~\citep{Guo_2022_CVPR} for the evaluation metrics. The evaluations are conducted 20 times to obtain a 95\% confidence interval. Methods indicated in \textit{italics} utilize the ground truth lengths for estimation. 
    Models above capable of bidirectional generation are also included in the captioning evaluation.
    For motion captioning, we use the ground truth captions without pre-processing and linguistic metrics suggested by \citet{guo2022tm2t} for evaluation.
    Best scores are highlighted in \textbf{boldface}, while \underline{underscore} refers to the second best. 
    }
    \vspace{-0.15in}
    \label{tab:t2m_table}
\end{table}

\subsection{Ablation Study}
\label{subsec:ablation}
\paragraph{Ablation on Motion-Agent}

\begin{figure}[h] 
\vspace{-0.1in}
\centering{
\includegraphics[width=\textwidth]{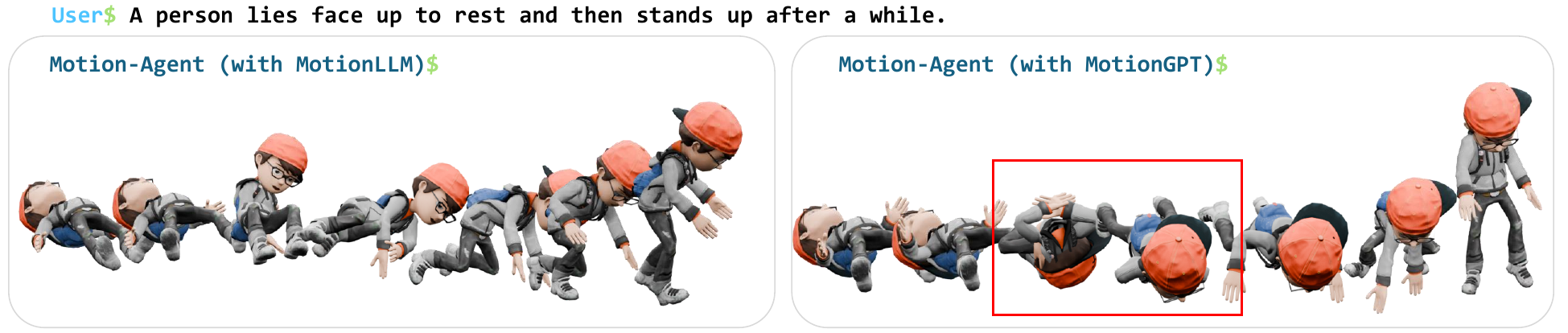}}
\vspace{-0.2in}
\caption{{\bf Motion-Agent Ablation Study}. We substituted MotionLLM with MotionGPT and noticed that MotionGPT cannot generate smooth motion transition.
}
\label{fig:ablation}
\end{figure}

Theoretically, the MotionLLM agent in our Motion-Agent framework can be replaced with any model capable of motion-text translation. However, models like MoMask~\citep{guo2023momask}, which require manual motion length input, may encounter issues (see Sec~\ref{subsubsec:qualitative_mllm}), making autoregressive models preferable. In this study, we substitute MotionLLM with MotionGPT~\citep{jiang2024motiongpt}, which also supports bidirectional translation. 
After integrating with Motion-Agent, we observe that MotionGPT is capable of generating longer and more complex motions compared to its original implementation. However, it still falls short of the accuracy and smoothness achieved by using MotionLLM. For example (Figure~\ref{fig:ablation}), in the user prompt ``A person lies face up to rest and then stands up after a while.'' GPT-4 decomposes this into two components: ``lying face up for a while'' and ``transition from lying face up to standing up.'' While MotionGPT correctly generates the first part, it incorrectly generates the second as ``from lying face down.'' This results in an abrupt and unsmooth transition between the two motions.
In contrast, MotionLLM accurately generates both parts, ensuring a smooth, seamless motion transition. 

Additionally, our framework can be adapted to use different LLMs for conversation. We tested substituting GPT-4 with various models, including Llama~\citep{touvron2023llama}, Gemma~\citep{team2024gemma2}, and Mixtral~\citep{jiang2024mixtral}. Most of these models successfully generated reasonable outputs and are capable of facilitating multi-turn interactions. Some smaller models may struggle with producing the correct JSON format. This ablation study demonstrated that our framework is applicable to all other LLMs, not just GPT-4. The performance of our framework can improve alongside the development of LLMs. More details are in Sec~\ref{subsec:different_llm}.

Nonetheless, our framework can be summarized as a combination of a larger LLM for conversation and a motion-language translation agent, providing flexible choices for different components.

\paragraph{Ablation on MotionLLM} 
We conducted an ablation study to examine the impact of different LLM backbones and adapter sizes. The results are shown in Table~\ref{tab:ablation}, from which we may conclude that using larger backbone models or increasing the LoRA rank leads to overall improvements in the metrics.

\begin{table}[h]
    \centering
    \resizebox{\textwidth}{!}{
    \begin{tabular}{ccccccc}
        \toprule
        \multirow{2}{*}{Models} & \multirow{2}{*}{Trainable Params} & \multicolumn{2}{c}{R Precision $\uparrow$} & \multirow{2}{*}{FID $\downarrow$} & \multirow{2}{*}{Multimodal Dist $\downarrow$} & \multirow{2}{*}{Diversity $\uparrow$}\\
        \cline{3-4}
        & & Top 1 & Top 3\\
        \midrule
        T2M-GPT~\citep{zhang2023generating} & 228.4M & 0.416 & 0.745 & 0.514 & 3.007 & 10.921 \\
         MotionGPT~\citep{jiang2024motiongpt} & 220M & 0.366 & 0.680 & 0.510 & 3.527 & 10.350 \\
        \midrule
        Gemma2-2b R=16 & 20.8M & 0.411 & 0.738 & 0.745 & 2.994 & 11.313\\
        Gemma2-2b R=32 & 41.5M & 0.415 & 0.750 & 0.712 & 2.938 & 11.251\\
        Gemma2-2b R=64 & 83.1M & 0.422 & 0.762 & 0.658 & 2.929 & 11.195\\
        LLaMA3-8B R=32 & 83.9M & 0.381 & 0.737 & 0.646 & 3.046 & 11.210\\
        Gemma2-9b R=32 & 108M & 0.439 & 0.776 & 0.438 & 2.872 & 11.151\\
        \bottomrule\\
    \end{tabular}
    }
    \vspace{-0.1in}
    \caption{\textbf{More comparisons and ablation study on the KIT-ML~\citep{Plappert2016} dataset}. Gemma~\citep{team2024gemma} and LLaMA~\citep{touvron2023llama} are chosen as LLM backbones. R indicates the LoRA rank, the value of alpha is kept the same with the rank. Two other autoregressvie transformer models are included for reference.
    }
    \vspace{-0.15in}
    \label{tab:ablation}
\end{table}

%% file: sections/5_discussion.tex
\section{Discussion}
\paragraph{Limitations and Future Work.} 
Our Motion-Agent specializes in generating motions of articulated 3D human body, without incorporating 3D visual understanding, such as interaction with the surrounding environment (e.g., ``a person puts his hand on the table''). Also, Motion-Agent does not include detailed hand or facial movements. Nonetheless, our framework demonstrates high flexibility, making it well-suited to incorporate additional agents for handling these tasks in future extensions. Additionally, while we have conducted preliminary trials on multi-human motion generation using our Motion-Agent framework—with some initial results (see Appendix~\ref{subsec:multi})—this has not yet been fully explored. Therefore, this paper still focuses on single-human motion generation. We left the extension for human-environment interaction and multi-human interaction for future work.

\paragraph{Concluding Remarks.} In this work, we propose a novel LLM-based multimodal, conversational motion-language learning framework, offering both flexibility and generalizability. By harnessing the linguistic comprehension and generation capabilities of pre-trained LLMs, our MotionLLM achieves strong results in bidirectional translation between motion and natural language. The Motion-Agent framework is easily expandable across various tasks through conversational interactions. Our approach is not only easy to train and adaptable but also user-friendly, making it a versatile solution for motion-language learning applications. Motion-Agent offers a comprehensive solution for enhancing LLMs' capabilities in understanding, generating, and editing human motion, aligning with our goal of teaching LLMs to interpret human motion effectively.

%% file: sections/appendix.tex
\newpage
\section{Appendix}

In the appendix, we present:
\begin{itemize}
    \item Section~\ref{subsec:qualitative_results}: More Qualitative Results.
    \item Section~\ref{subsec:different_llm}: Ablation study on different LLMs
    \item Section~\ref{subsec:multi}: Preliminary Trials on Multi-human Motion Generation using Motion-Agent
    \item Section~\ref{subsec:eval_metric}: More details about the evaluation metrics.
    \item Section~\ref{subsec:moreImplementDetails}: More details regarding our implementation.
\end{itemize}

\subsection{Qualitative Results}
Rendered original videos of all examples shown in the paper can be found in the corresponding folder of the supplementary material.
\label{subsec:qualitative_results}
\subsubsection{Motion-Agent}
\label{subsubsec:moton-agent-qualitative}
More examples of Motion-Agent are presented in Figure~\ref{fig:demo4}, \ref{fig:demo2}, \ref{fig:demo3}, and corresponding videos can be found in the supplementary material.
\begin{figure}[ht]
    \centering
    \includegraphics[width=\textwidth]{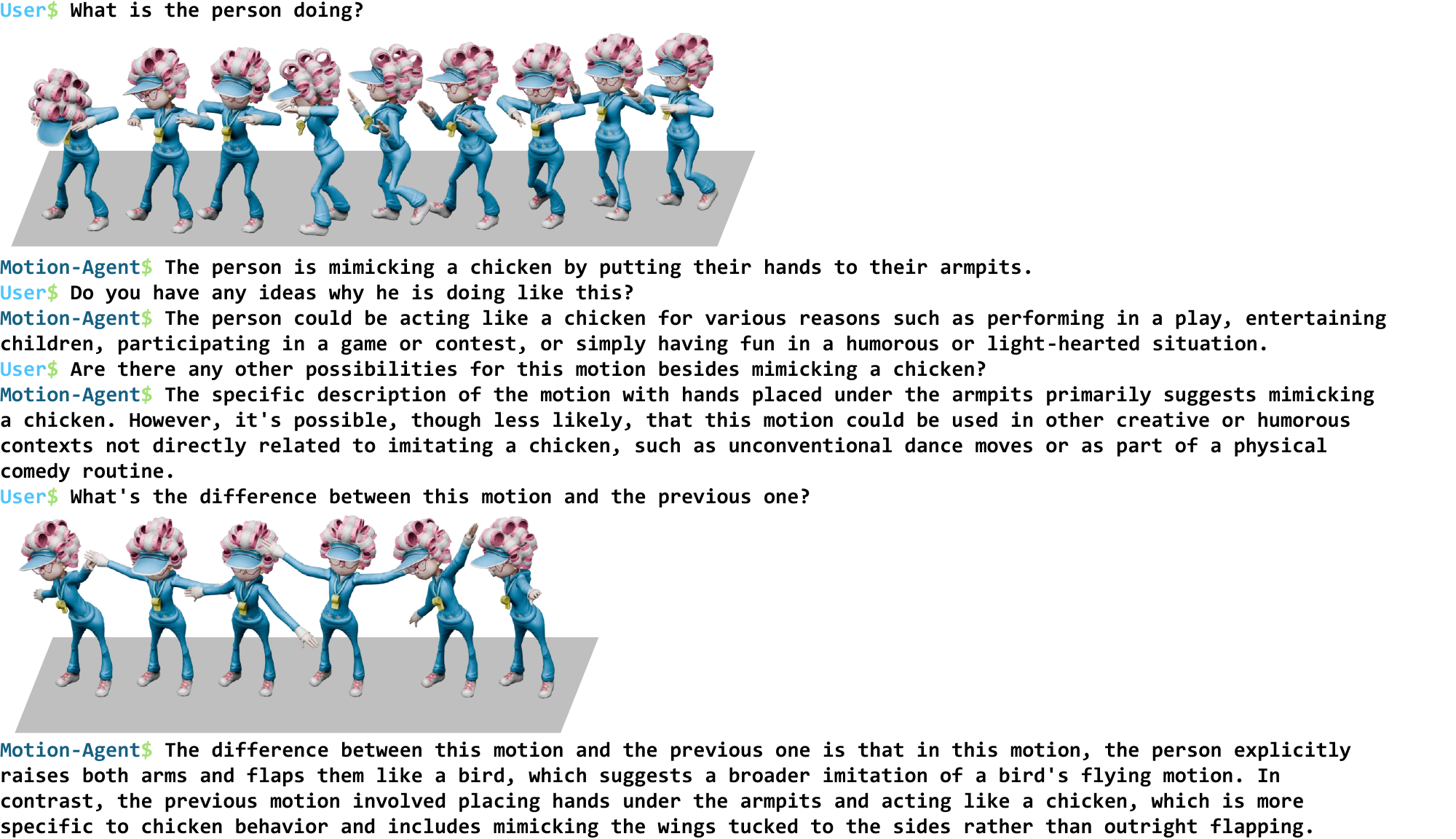}
    \caption{More examples of Motion-Agent.}
    \label{fig:demo4}
\end{figure}
\begin{figure}[ht]
    \centering
    \includegraphics[width=\textwidth]{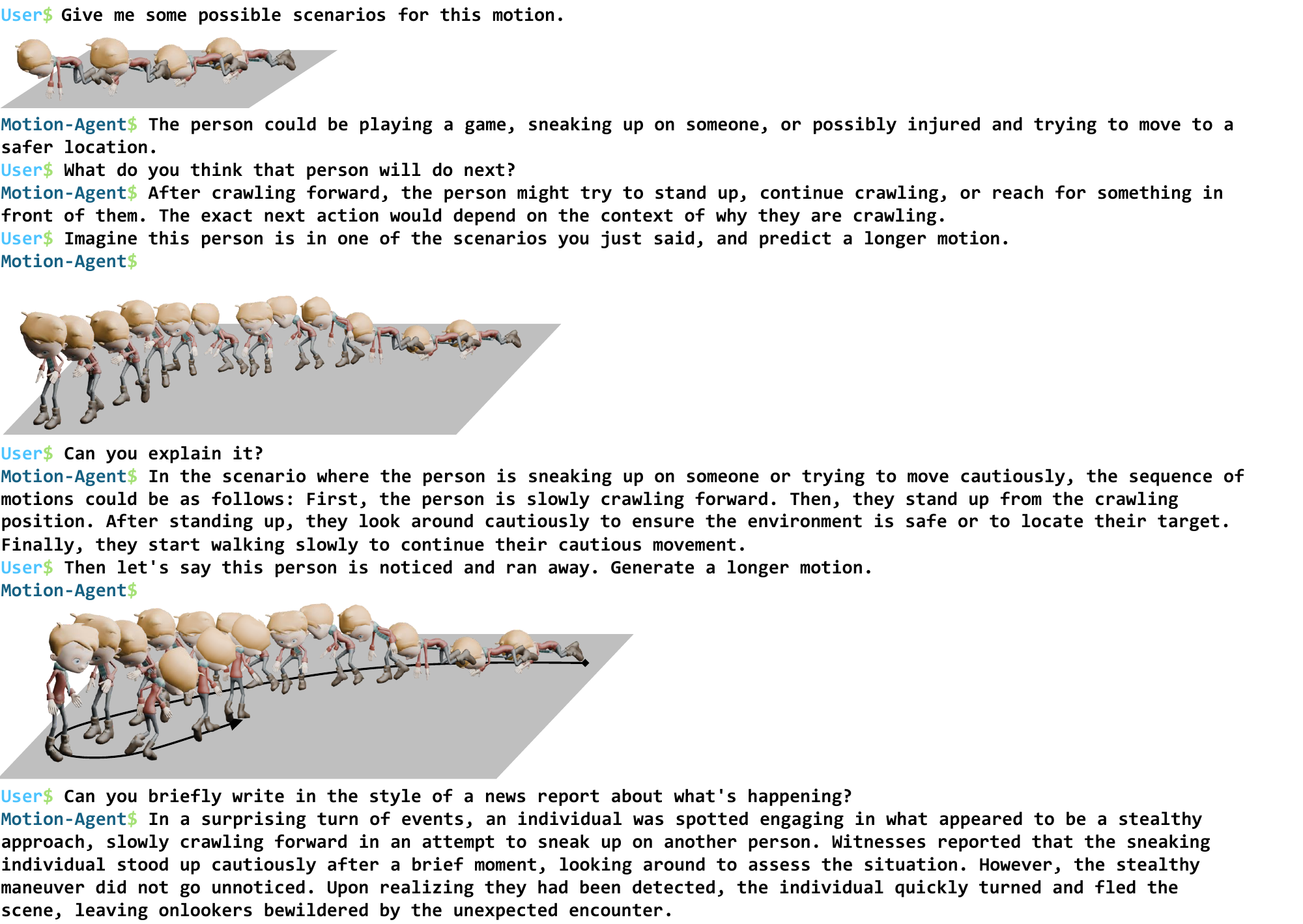}
    \caption{More examples of Motion-Agent.}
    \label{fig:demo2}
\end{figure}
\begin{figure}[ht]
    \centering
    \includegraphics[width=\textwidth]{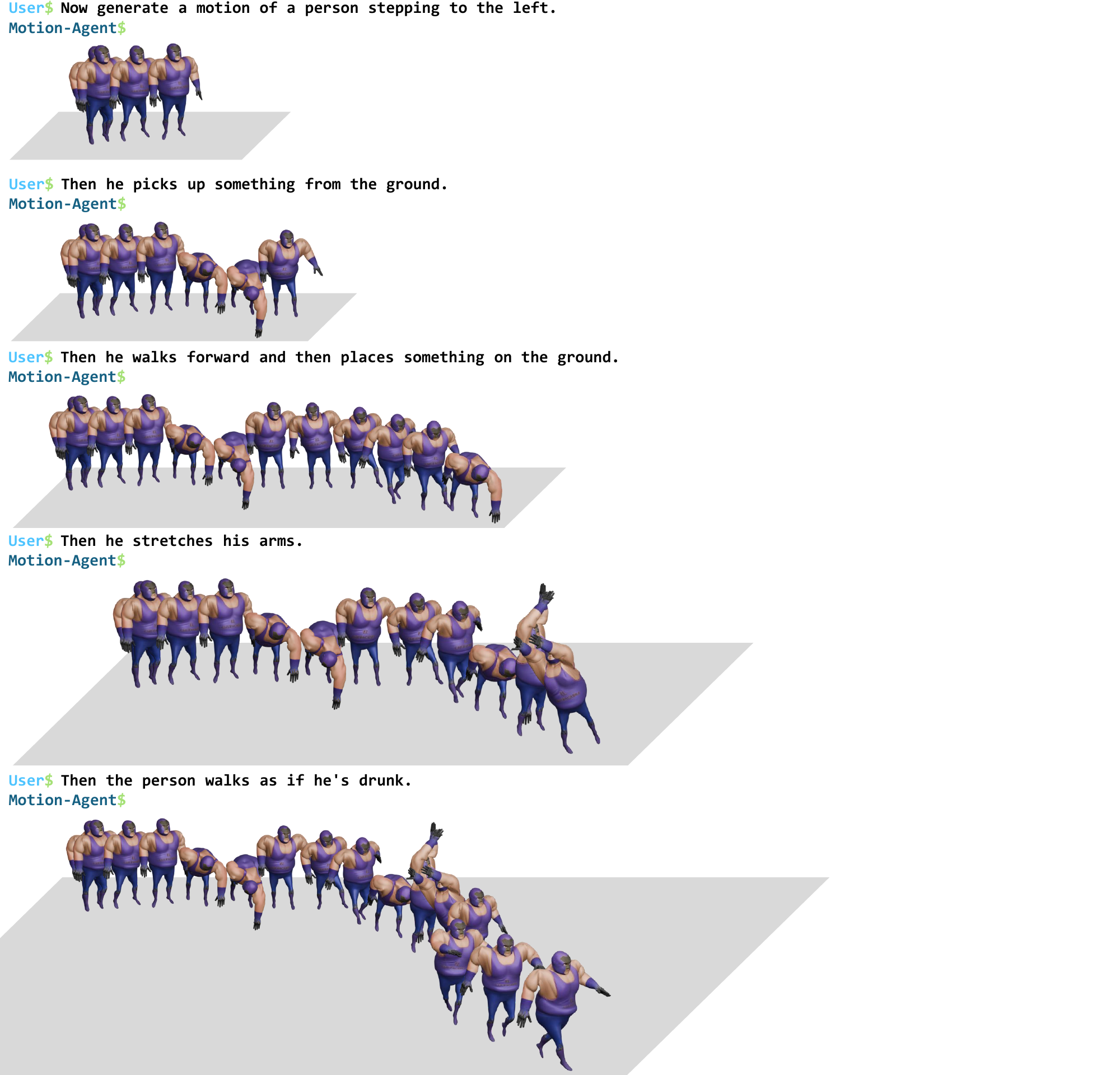}
    \caption{More examples of Motion-Agent.}
    \label{fig:demo3}
\end{figure}

\subsubsection{MotionLLM}
\label{subsubsec:qualitative_mllm}
\begin{figure}[ht]
    \centering
    \includegraphics[width=\textwidth]{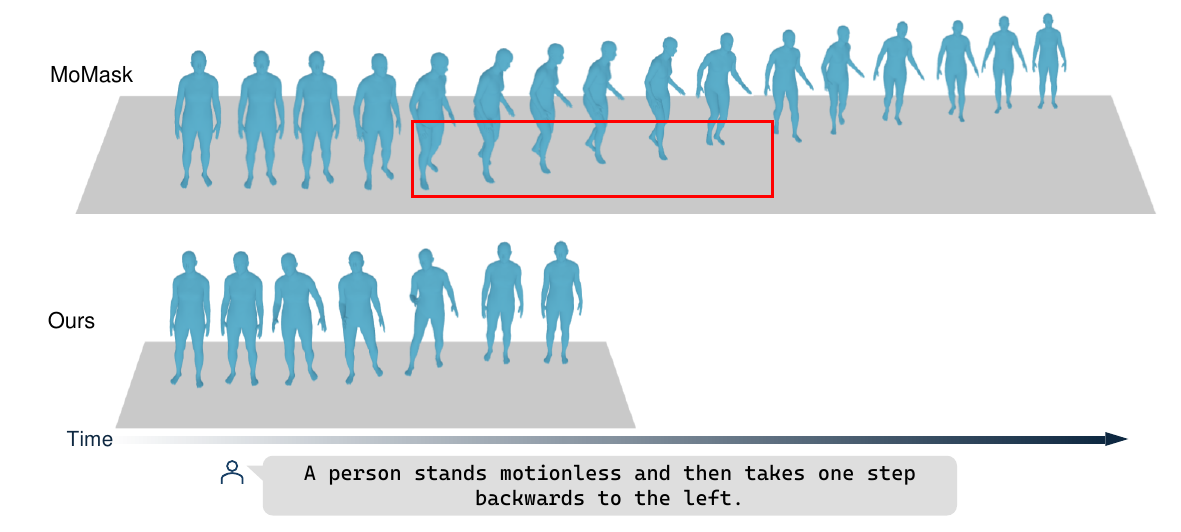}
    \caption{Comparison between MotionLLM and MoMask~\citep{guo2023momask}, which is non-autoregressive. During regular inference, MoMask uses a length estimator to predict the length conditioned on the text. This estimator is likely to fail. In this example, their incorrect predicted length causes severe drifting.}
    \label{fig:vs_guo}
\end{figure}

\paragraph{Motion Generation}
Figure~\ref{fig:vs_guo} presents the comparison on no-length-given motion generation. More qualitative results are in the supplementary material.

\paragraph{Motion Captioning}

\begin{table}[ht]
\resizebox{\textwidth}{!}{
\begin{tabular}{ccl}
\toprule
Motion  & Model   & \multicolumn{1}{c}{Caption}  \\ \hline
& \cellcolor[HTML]{EFEFEF}Ground Truth & \cellcolor[HTML]{EFEFEF}a person walks forward just like a mummy                                                                        \\
& TM2T                                 & a person walk in a counterclockwise circle with their arm out to the side                                                              \\
& \cellcolor[HTML]{EFEFEF}MotionGPT    & \cellcolor[HTML]{EFEFEF}the person is walking like a mummy from the dead                                                               \\
\multirow{-4}{*}{\begin{tabular}[c]{@{}c@{}}demo\_1\end{tabular}} 
& Ours                                 & a person walks forward while holding arms out as if to be a zombie                                                                     \\ \hline
& \cellcolor[HTML]{EFEFEF}Ground Truth & \cellcolor[HTML]{EFEFEF}a person walks forward slowly while their right hand is slightly elevate                                        \\
& TM2T                                 & a person slowly walk forward while hold onto something with their left hand                                                            \\
& \cellcolor[HTML]{EFEFEF}MotionGPT    & \cellcolor[HTML]{EFEFEF}\begin{tabular}[c]{@{}l@{}}a person walks forward slowly, placing one foot in front of the other, on a belt \\ that circulates, enabling the person to effectively slowly walk in place.\end{tabular}                                                                                                                                  \\
\multirow{-4}{*}{\begin{tabular}[c]{@{}c@{}}demo\_2\end{tabular}} 
& Ours                                 & the person is walking on a balance beam                                                                                                \\ \hline
& \cellcolor[HTML]{EFEFEF}Ground Truth & \cellcolor[HTML]{EFEFEF}a person moves side to side in a zigzag fashion backwards                                                       \\
& TM2T                                 & a person does a cartwheel to the right                                                                                                 \\
& \cellcolor[HTML]{EFEFEF}MotionGPT    & \cellcolor[HTML]{EFEFEF}a person is practing defense moves.                                                                            \\
\multirow{-4}{*}{\begin{tabular}[c]{@{}c@{}}demo\_3\end{tabular}}  
& Ours                                 & a person walks backwards in zig-zag motion                                                                                             \\ \hline
& \cellcolor[HTML]{EFEFEF}Ground Truth & \cellcolor[HTML]{EFEFEF}a person makes and drinks a cup of coffee                                                                        \\
& TM2T                                 & \begin{tabular}[c]{@{}l@{}}person hold something with their right hand and make a sawing motion with \\ their left hand\end{tabular}   \\
& \cellcolor[HTML]{EFEFEF}MotionGPT    & \cellcolor[HTML]{EFEFEF}a person is eating something                                                                                   \\
\multirow{-4}{*}{\begin{tabular}[c]{@{}c@{}}demo\_4\end{tabular}} 
& Ours                                 & \begin{tabular}[c]{@{}l@{}}a person uses their left hand to open a bottle, drinks from it, then places the \\ bottle back down\end{tabular}\\                              
\bottomrule
\end{tabular}
}
\caption{Comparsion of motion captioning ability across different models. Original motions can be found in supplementary material.}
\label{table:mocap compare}
\end{table}

Our model is capable of generating high-quality motion captions, demonstrations are provided in Table \ref{table:mocap compare}.

\subsection{Ablation Study on Different LLMs} 
\label{subsec:different_llm} 
In this study, we replace GPT-4 with several other LLMs, including Llama~\citep{touvron2023llama}, Gemma~\citep{team2024gemma2}, and Mixtral~\citep{jiang2024mixtral}. The experiment involved a straightforward two-turn conversation. In the first turn, the we prompted, "Generate a motion that a person is doing exercise." In the second turn, we provided a motion that a person is slowly crawling forward and asked, "Briefly explain the possible scenarios for this motion." The decomposed arguments from the agent in the first turn and the response from the second turn are presented in Table~\ref{tab:llm_ablatioin}.

Overall, we observe that different LLMs can generate reasonable outputs in response to user requests. However, smaller models, such as Llama-3-7B and Mixtral-8x7B, while capable of producing some acceptable responses, struggled to adhere strictly to the instructed JSON format. As a result, the agent was unable to parse their outputs successfully.

\begin{table}[!ht]
\centering
\renewcommand{\arraystretch}{1.5}
\begin{tabular}{p{2cm}p{5cm}p{5cm}}
\hline
\textbf{LLM} & \textbf{Turn 1 Decomposition Results}                            & \textbf{Turn 2 Reasoning Response}                            \\ \hline
GPT-4-Turbo        & \textit{'A person does jumping jacks.', 'A person does push-ups.', 'A person does sit-ups.', 'A person stretches.'} & \textit{"The person could be participating in a military training exercise, playing a game that involves crawling, navigating through a confined space, or engaging in a fitness routine that includes floor-based movements."} \\ \hline
Llama-3-70B                       & \textit{'A person stands up straight.', 'A person raises their arms above their head.', 'A person bends their knees slightly.', 'A person starts jogging in place.', 'A person lowers their arms to their sides.'}   & \textit{"The person could be participating in a military or law enforcement training exercise, engaging in a crawling competition, playing a game that involves crawling, or navigating through a confined or obstacle-filled space."}                                \\ \hline
Gemma2-27B & \textit{'A person does jumping jacks.', 'A person does push-ups.', 'A person does sit-ups.'} & Possible scenarios include: army training exercise, searching for something lost on the ground, playing a game, navigating a tight space, or enacting a dramatic scene.\\ \hline
Llama-3.1-8B                       & \textit{'A person starts in a standing position.', 'A person bends their knees.', 'A person lowers their body.', 'A person stands up.', 'A person repeats the motion.'}        & \textit{"The person could be crawling through a tight space, crawling on the floor for exercise, or crawling as a form of slow and deliberate movement in a situation where walking or running is not feasible."}               \\ \hline
Llama-3-7B & Failed to follow the format & Failed to follow the format \\ \hline
Mixtral-8x7B & Failed to follow the format & Failed to follow the format \\ \hline
\end{tabular}
\caption{Comparison of Using Different LLMs on a Two-Turn Prompt.}
\label{tab:llm_ablatioin}
\end{table}

\subsection{Multi-human with Motion-Agent}
\label{subsec:multi}

\begin{figure}[ht]
    \centering
    \includegraphics[width=\textwidth]{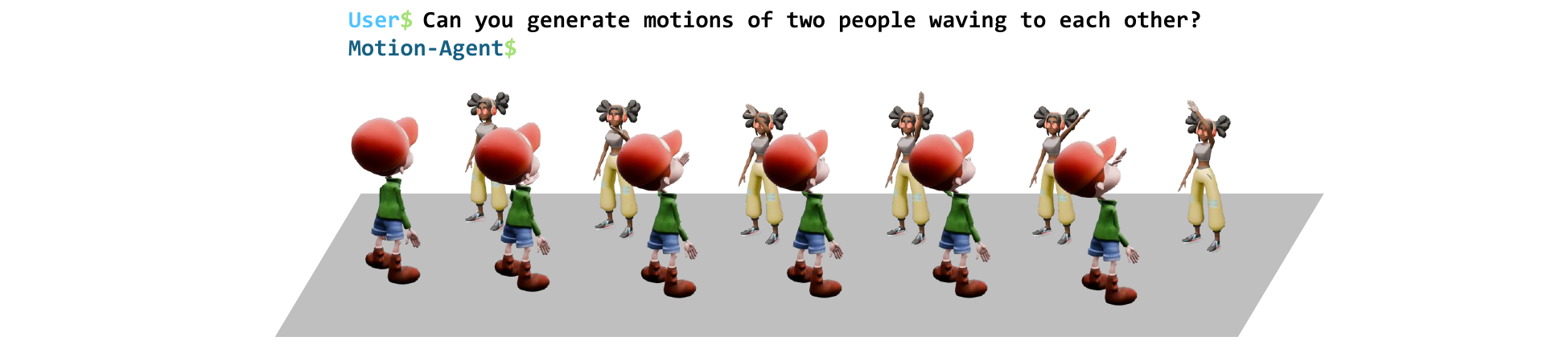}
    \caption{Multi-human Motion Generation using Motion-Agent.}
    \label{fig:multi}
\end{figure}
In this section, we present the results of our preliminary trials on multi-human motion generation using the Motion-Agent framework, specifically focusing on generating motions for two individuals.

In our implementation, each person is represented in the HumanML format~\citep{Guo_2022_CVPR}, with their motions defined separately. To uniquely define the motions of both individuals, we incorporate location information for the first frame, represented by a tuple of three parameters, relative $r=(\theta, x, z)$. Here, the first person is always positioned at the origin in 3D space, and the relative tuple $r$ determines the position of the second person concerning the first. The parameter $\theta$ denotes the rotation radius, while $x$ and $z$ represent the coordinates (with the y-axis as vertical). Therefore, the motion of each person together with $r$ can uniquely determine the whole motion. In this context, GPT-4 is tasked with generating three outputs: the arguments for MotionLLM for each person and the relative tuple $r$.

Figure~\ref{fig:multi} shows an example of multi-human generation using Motion-Agent. In this case, the two arguments are "A person waves.", and $r=(3.14,0,1)$, indicating that the second person rotates 180 degrees (since $3.14\approx \pi$) from facing the $z^{+}$ direction (hence positioned face to face with the first person) and is standing 1 meter away from the first person.
\subsection{Evaluation Metric}
\label{subsec:eval_metric}
We detail the calculation of several evaluation metrics proposed in \cite{Guo_2022_CVPR}. We denote ground-truth motion features, generated motion features, and text features as $f_{\text{gt}}$, $f_{\text{pred}}$, and $f_{\text{text}}$. Note that these features are extracted with pretrained networks in \cite{Guo_2022_CVPR}.

\textbf{Multimodal Distance (MM-Dist).} MM-Dist is widely used to evaluate the motion generation ability of the model. MM-Dist measures the distance between the text embedding and the generated motion feature.  Given $N$ randomly generated samples, the MM-Dist measures the feature-level distance between the motion and the text. It computes the average Euclidean distances between each text feature and the generated motion feature from this text:
\[
\text{MM-Dist} = \frac{1}{N} \sum_{i=1}^N \|f_{\text{pred},i} - f_{\text{text},i}\|
\]
where $f_{\text{pred},i}$ and $f_{\text{text},i}$ are the features of the $i$-th text-motion pair.

\textbf{Frechet Inception Distance (FID).} FID measures the distance of motion features distribution between real and generated motions. We calculate FID by
\[
\mathit{FID} = \| \mu_{\text{gt}} - \mu_{\text{pred}} \|^2 - \text{Tr}(\Sigma_{\text{gt}} + \Sigma_{\text{pred}} - 2(\Sigma_{\text{gt}} \Sigma_{\text{pred}})^{1/2})\]
where $\mu_{\text{gt}}$ and $\mu_{\text{pred}}$ are the means of $f_{\text{gt}}$ and $f_{\text{pred}}$. $\Sigma$ is the covariance matrix and $\operatorname{Tr}$ denotes the trace of a matrix.

\textbf{R precision} Given the motion sequence and 32 text descriptions (1 ground-truth and 31 randomly selected mismatched descriptions), we rank the Euclidean distances between the motion and text embeddings to get Top-1, Top-2, and Top-3 accuracy of motion-text;

\textbf{Diversity.} Diversity measures the variance of the whole motion sequences across the dataset. We randomly sample $S_{\text{dis}}$ pairs of motion and each pair of motion features is denoted by $f_{\text{pred},i}$ and $f'_{\text{pred},i}$. The diversity can be calculated by
\[
\mathit{Diversity} = \frac{1}{S_{\text{dis}}} \sum_{i=1}^{S_{\text{dis}}} \|f_{\text{pred},i} - f'_{\text{pred},i}\|
\]
In our experiments, we set $S_{\text{dis}}$ to 300 as ~\citep{Guo_2022_CVPR}.

\textbf{Linguistic
 metrics.} Linguistic
 metrics including Bleu~\citep{bleu}, Rouge~\citep{lin-2004-rouge}, Cider~\citep{vedantam2015cider} and Bert Score~\citep{zhang2020bertscore}, we follow TM2T~\citep{guo2022tm2t}, using NLPEval to calculate. Readers can
 refer to their papers for further details.

\subsection{More Implementation details}
\label{subsec:moreImplementDetails}

\paragraph{Prompts For MotionLLM.} We use different prompts for different tasks.

\begin{table}[H]
    \centering
    \begin{tabular}{cc}
        \toprule
        Task & Prompts\\
        \hline
        Motion Generation & \makecell{Generate a motion matching the following\\ input human motion description.}\\
        \hline
        Motion Captioning & \makecell{Generate a caption matching the following\\ input human motion token sequence.}\\
        \bottomrule\\
    \end{tabular}
    \caption{Instructing prompts for MotionLLM training and inference.}
    \label{tab:prompts}
\end{table}
\paragraph{Hyper-parameters.} Our hyper-parameters settings for different tasks.
\begin{table}[H]
    \centering
    \begin{tabular}{ccc}
        \toprule
        Hyper-parameter & Motion Generation & Motion Captioning \\
        \midrule
        Batch size & 6 & 6\\
        Learning rate & 1e-5 & 1e-5\\
        LoRA rank & 64 & 32\\
        LoRA alpha & 32 & 32\\
        LoRA dropout & 0.1 & 0.1\\
        Codebook size & 512 & 512\\
        Codebook dim & 512 & 512\\
        Total vocab size & 256514 & 256514\\
        \bottomrule\\
    \end{tabular}
    \caption{Hyper-parameters of our models used in our main experiments. Other VQ training settings are borrowed from T2M-GPT~\citep{zhang2023generating}}
    \label{tab:hyper-params}
\end{table}